\documentclass[12pt]{article}
\makeatletter

\usepackage{amsmath,amssymb,amsfonts}
\usepackage{algorithmic}
\usepackage{textcomp}
\usepackage{xcolor}
\usepackage{cite}
\usepackage{booktabs} 
\usepackage{url}%
\usepackage[top=0.75in,bottom=1in,left=0.75in,right=0.75in,columnsep=0.2in]{geometry} 

\usepackage{blindtext}
\usepackage{eso-pic}
\usepackage{graphicx}
\usepackage{textcomp}
\usepackage{eso-pic}

\begin{document}

\title{A Lightweight Federated Learning Approach for Privacy-Preserving Botnet Detection in IoT}

\author{Taha M. Mahmoud \\ University of North Dakota \\ \texttt{taha.mahmoud@und.edu} 
\and Naima Kaabouch \\ University of North Dakota \\ \texttt{naima.kaabouch@und.edu}}
\date{}

\maketitle

\begin{center}
\textit{This work has been published in the Proceedings of the 2025 IEEE International Conference on Applied Cloud and Data Science and Applications (ACDSA). 
The final published version is available via IEEE Xplore at \url{https://doi.org/10.1109/ACDSA65407.2025.11165820}.}
\end{center}

\begin{abstract} The rapid growth of the Internet of Things (IoT) has expanded opportunities for innovation but also increased exposure to botnet-driven cyberattacks. Conventional detection methods often struggle with scalability, privacy, and adaptability in resource-constrained IoT environments. To address these challenges, we present a lightweight and privacy-preserving botnet detection framework based on federated learning. This approach enables distributed devices to collaboratively train models without exchanging raw data, thus maintaining user privacy while preserving detection accuracy. A communication-efficient aggregation strategy is introduced to reduce overhead, ensuring suitability for constrained IoT networks. Experiments on benchmark IoT botnet datasets demonstrate that the framework achieves high detection accuracy while substantially reducing communication costs. These findings highlight federated learning as a practical path toward scalable, secure, and privacy-aware intrusion detection for IoT ecosystems. \end{abstract}

\textbf{Key Words:} IoT Security, Botnet Detection, Federated Learning, Lightweight Machine Learning, Decision Tree, Anomaly Detection

\maketitle

\pagenumbering{gobble}

\section{Introduction} \label{intro}

The Internet of Things (IoT) has seamlessly integrated into our daily lives, powering everything from smart homes to healthcare and transportation systems. However, as the number of interconnected devices continues to grow, so do the associated security challenges. Traditional security mechanisms often fall short, especially against sophisticated botnet attacks that can simultaneously exploit numerous devices. Detecting these attacks is particularly challenging due to the diverse nature of IoT devices and their inherent limitations in computational power, memory, and energy resources \cite{kambourakis_mirai_2017}

Botnet attacks have long been a significant concern in cybersecurity. Notably, the 2016 Mirai botnet attack, which compromised over 600,000 IoT devices by exploiting weak or default credentials, caused widespread internet service disruptions. Similarly, the 2014 Bashlite botnet attack targeted vulnerabilities in Linux-based systems to commandeer IoT devices for launching DDoS attacks. These incidents underscore the urgent need for robust security measures in IoT environments.

Existing solutions for botnet detection in IoT networks fail to effectively balance the need for accurate detection with the constraints of privacy, computational power, and resource availability on IoT devices. The reliance on centralized data processing not only increases vulnerability to data breaches but also limits the scalability and adaptability of these solutions in diverse and evolving IoT landscapes. Moreover, lightweight machine learning models, which are suitable for resource-constrained devices, have not been fully leveraged within a decentralized, privacy-preserving framework to address these challenges.

Therefore, there is a critical need for a new botnet detection framework that optimally balances high detection accuracy with minimal computational overhead while preserving data privacy. This study seeks to fill this gap by investigating the use of federated learning combined with lightweight machine learning algorithms to develop a decentralized, privacy-preserving, and scalable botnet detection framework for IoT ecosystems. By doing so, the research aims to provide a more effective and secure method for combating botnet threats, addressing both the technological constraints and privacy concerns inherent to IoT environments.

The primary objective of this study is to develop a botnet detection framework that is both robust and privacy-preserving, specifically designed for the constrained environments of IoT ecosystems. To achieve this, we aim to identify the most effective lightweight supervised machine learning algorithm—such as decision trees, k-nearest neighbors (KNN), support vector machines (SVM), or logistic regression—that optimally balances high detection accuracy with low training time. This selection process focuses on algorithms that can operate efficiently on IoT devices with limited computational resources, including restricted CPU, RAM, and storage capacities.

Another key objective is to implement a federated learning approach, where multiple IoT devices collaboratively train a model using their local data without data exchange. This approach aims to enhance privacy and security by design, making the detection process more resilient against data breaches. We will evaluate the performance of the decentralized federated learning model against traditional centralized learning methods using the N-BaIoT dataset, focusing on improvements in security, privacy, and resource efficiency.

This study introduces a significant advancement in botnet detection by employing an ensembling technique on locally trained lightweight classifier models to simulate federated learning. By combining the predictions of multiple locally trained models, the ensemble approach captures a wider range of patterns, enhancing the generalization and robustness of the detection framework. This technique also increases scalability by enabling independent model training on numerous IoT devices without requiring data sharing. The proposed framework leverages the strengths of classical machine learning models, achieving effective botnet detection while maintaining data privacy and ensuring scalable deployment across diverse IoT environments.

The paper begins with an \textbf{Introduction} \ref{intro}, which explains the rapid growth of IoT devices and the increasing security challenges they face. In the \textbf{Related Work} \ref{RW}, the authors explore previous work related to IoT security and federated learning. The \textbf{Methodology} \ref{Methodology} section describes how the experiments were designed and how the data was handled. In the \textbf{Results and Discussion} \ref{RD}, the paper presents detailed findings supported by both numbers and observations. Then explains what these results mean. Finally, the \textbf{Conclusion} \ref{conc} summarizes the key outcomes and suggests future research directions to tackle remaining challenges.

\section{Related Work}\label{RW}
The current work focuses on various methods for detecting botnet activities in IoT networks, an increasingly critical issue due to the proliferation of IoT devices and their inherent security vulnerabilities. The key methods explored are \textbf{statistical analysis}, \textbf{machine learning supervised methods}, \textbf{unsupervised clustering methods}, and \textbf{federated learning approaches}.

\textbf{Statistical Analysis}: This method, which uses basic traffic metrics like packet sizes and connection intervals, is lightweight and suitable for real-time detection. However, it cannot identify complex, multi-dimensional attacks and has minimal integration with privacy-preserving frameworks like federated learning. Techniques like Extreme Value Theory \cite{siffer2017anomaly} and ARIMA models \cite{kozitsin2021online} have been proposed, showing promise for time-series anomaly detection, but their univariate nature limits their broader applicability. These models are particularly weak in detecting botnet pre-attack stages and handling multi-feature detection, requiring separate models for each dimension, which can result in loss of correlation between features. Additionally, statistical methods may struggle with detecting sophisticated attacks like zero-day threats or multi-stage botnet activities.

\textbf{Machine Learning Supervised Methods}: These methods, including Support Vector Machines (SVMs) and deep learning models like autoencoders \cite{Meidan2018} and Recurrent Neural Networks (RNNs) \cite{botnet2018deeplearning}, offer high accuracy but depend heavily on the quality of labeled training data. While models such as deep autoencoders \cite{Meidan2018} and bidirectional LSTM-RNNs \cite{botnet2018deeplearning} have been effective in identifying IoT botnet activities, the reliance on manually labeled data is a major limitation, as labeled data may not always be available, particularly for new or evolving attacks. The computational cost of training these models is another concern, especially in resource-constrained IoT environments, and training times may be further prolonged when integrated with federated learning for privacy protection. Additionally, feature selection techniques such as decision trees and principal component analysis have shown improvements in model accuracy, but their integration with neural networks \cite{alauthaman2018p2p} still requires further refinement.

\textbf{Unsupervised Clustering Methods}: Techniques like K-means and DBSCAN are useful for detecting unknown or zero-day attacks by clustering similar traffic patterns. Although they do not require labeled data, these methods tend to have lower accuracy and higher false positive rates, making them less reliable for real-time botnet detection. The lack of labeled data complicates the evaluation of the model’s performance, and clustering approaches can be sensitive to the choice of hyperparameters, leading to inconsistent results. Moreover, integrating these methods with federated learning may be challenging due to the difficulty in coordinating clusters across distributed nodes. Despite their ability to identify evolving botnet behaviors, the higher false positive rates are a significant drawback \cite{guo_review_2023}, \cite{zoppi_unsupervised_2021}.

\textbf{Federated Learning}: Federated learning is emerging as a promising solution for botnet detection, enabling collaborative model training across IoT devices while maintaining data privacy. Methods like Federated Deep Learning (FDL) \cite{popoola2021federated} and K-greedy aggregation \cite{10279423ZeroDayBotnet} have demonstrated improved accuracy in detecting zero-day attacks. However, federated approaches often suffer from extended training times and potential vulnerability to poisoning attacks, where compromised devices can influence the global model's performance \cite{famera2023cross}. Furthermore, federated learning requires careful management of computational resources across IoT devices, which are often resource-constrained, and the communication overhead between devices can further complicate real-time detection. Despite these challenges, federated learning offers a promising framework for privacy-preserving botnet detection, as shown in recent work \cite{regan2022federated}.

The research questions guiding this project are designed to investigate the performance and efficiency of various machine-learning techniques for botnet detection within IoT environments. Specifically, these techniques will be compared based on accuracy and training time, the most resource-efficient methods suitable for IoT devices will be identified, and the effectiveness of federated learning  technologies in detecting anomalies will be assessed. Additionally, it will explore how these technologies can be integrated to create a real-time, secure, and private botnet detection framework. This inquiry aims to comprehensively understand the potential synergies between machine learning, and federated learning  in enhancing IoT security.

\section{Methodology}\label{Methodology}
The main goal of this project is to develop an efficient machine-learning model that can balance the required time to train the model and its accuracy. This model will be deployed on IoT devices with limited resources regarding compute power, RAM, storage, and network bandwidth. To achieve this goal, a framework based on federated machine learning  will be introduced to showcase how this technique can construct a powerful platform that preserves privacy and security simultaneously.

\begin{figure}
     \centering
\includegraphics[width=0.75\linewidth]{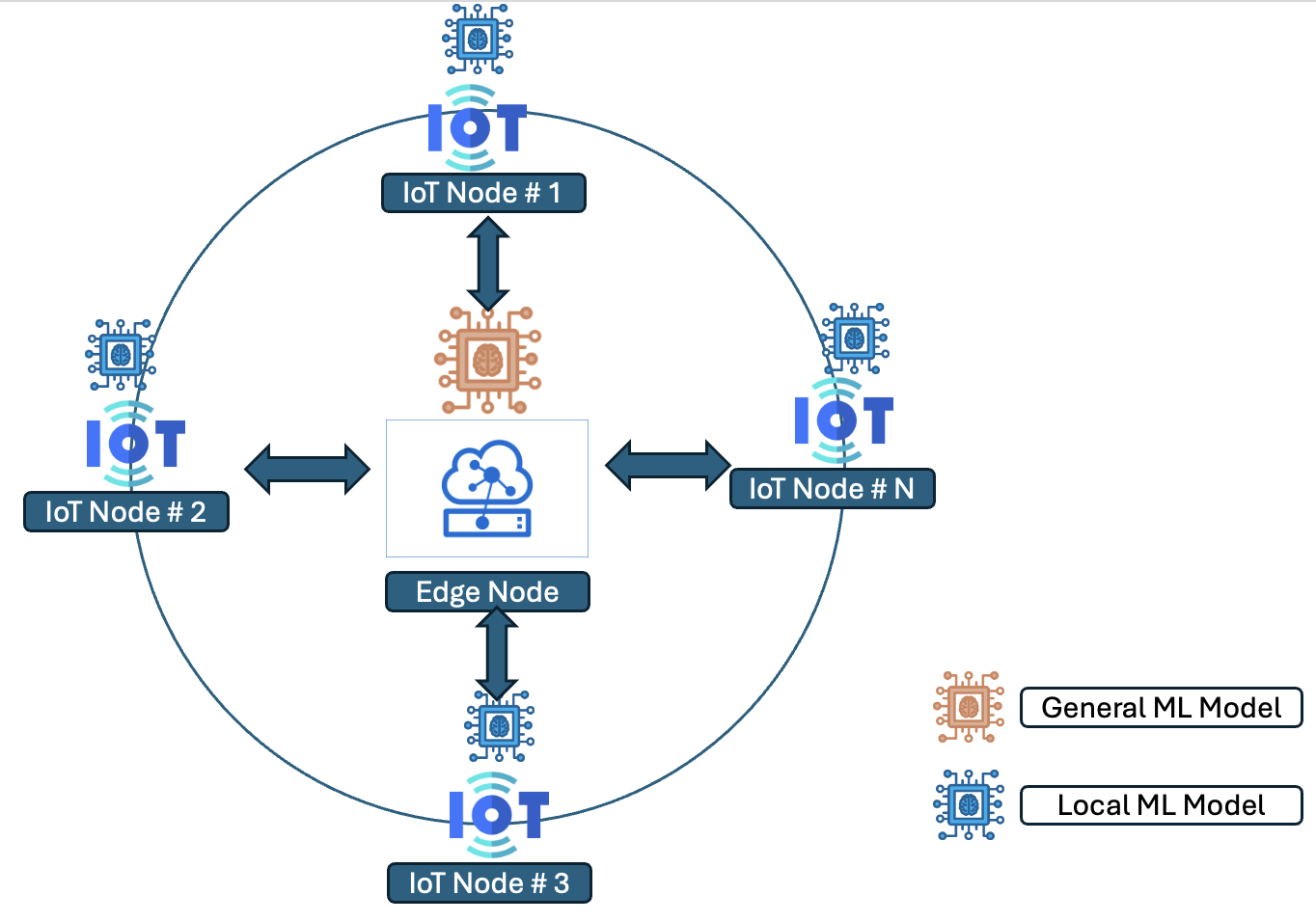}
     \caption{Botnet detection framework}
     \label{fig:botnet_detection_framework}
 \end{figure}

Figure \ref{fig:botnet_detection_framework} illustrates a botnet detection overall system model for IoT networks based on federated learning, featuring two main components: IoT Nodes and an Edge Node. Each IoT Node represents an individual IoT device that locally trains a machine learning model, referred to as the "Local ML Model," using its own data to detect potential botnet activities. The local models help identify anomalies or suspicious behavior directly on the devices, maintaining data privacy and reducing the need for raw data transmission.

The edge node serves as a central point in the network that brings together updates from all IoT devices. Instead of collecting raw data, it gathers only the trained model updates from each device. It then combines these updates to create a refined global model, called the “General ML Model.” This improved model is shared back with the devices, helping them better detect threats based on a broader understanding learned from across the network.

This framework ensures a collaborative learning process where IoT devices contribute to and benefit from a shared model, while also preserving data privacy and minimizing network bandwidth usage. The iterative cycle of local training, model aggregation, and redistribution depicted in the diagram forms the core methodology for achieving efficient and scalable botnet detection across distributed IoT environments.

The experiments were conducted using both a MacBook Air with an Apple M1 chip (16 GB RAM, macOS Ventura) and Google Colab, depending on the dataset size and processing requirements. Python 3.10 was used along with common libraries including pandas, scikit-learn, NumPy, and matplotlib. No deep learning frameworks or GPU acceleration were used, to better reflect deployment in constrained IoT environments.

The N-BaIoT dataset was used to train and evaluate machine learning models for botnet detection. It contains labeled network traffic data from nine IoT devices infected by Mirai and BASHLITE botnets, covering one normal communication type and ten attack types, with about 115 features per instance. Due to missing data in devices 3 and 7, the analysis focused on the remaining seven devices, especially devices 1 and 4, which had the most complete data. The dataset is well-organized and required minimal preprocessing. The attack files were merged and labeled, and the data was split into training and testing sets (80\%, and 20\%). 

For botnet detection in IoT networks, lightweight machine learning models such as Decision Trees, SVM, KNN, and Logistic Regression are well-suited due to their low computational requirements and ability to operate in real-time environments. These models are ideal for resource-constrained IoT devices and can be integrated into federated learning frameworks, allowing decentralized training while preserving data privacy.

\begin{itemize}
    \item \textbf{Decision Trees} are fast and interpretable, making them suitable for real-time botnet detection, though they may struggle with complex patterns without proper pruning.
    \item \textbf{SVM}, particularly with linear kernels, is effective for binary classification tasks like botnet detection but may become computationally expensive on large datasets, especially with non-linear kernels.
    \item \textbf{KNN} is easy to implement and performs well in real-time, but its computational cost increases with larger datasets, affecting real-time performance.
    \item \textbf{Logistic Regression} is efficient for binary classification and performs well on large datasets, particularly when relationships are linear.
\end{itemize}

\textbf{Scaling} is crucial for distance-based algorithms like KNN and SVM to avoid bias caused by differing feature ranges. While \textbf{PCA} was expected to reduce training time for KNN and SVM, it was not used due to concerns about impacting model accuracy, especially when applied to data from another IoT node or in federated learning, as PCA’s transformation could distort feature relationships across nodes.

Given their lightweight nature and ability to be integrated into federated learning setups, these models offer a good balance between computational efficiency and detection accuracy. They can be quickly trained on local IoT devices, with the aggregated global model refined iteratively to improve detection performance across the network. While their performance in terms of accuracy and training time may vary based on dataset size and complexity, these models are generally expected to deliver reliable results in real-time botnet detection scenarios, with Decision Trees and Logistic Regression favoring faster training times and SVM and KNN potentially offering higher accuracy with careful tuning.

\textbf{Federated Machine Learning}
   We will conduct an experiment to show that while machine learning models tend to perform well on the IoT node dataset used for training, they may not perform well on other IoT node datasets. For example, a model trained on Node 1 dataset may perform well on testing datasets from that node but not on IoT node 2 training data. This highlights the importance of introducing federated machine learning techniques that share insights and knowledge between IoT nodes while maintaining each node's privacy by not sharing any local data. 

 Our novel methodology uses ensemble methods to build a general model at the edge node based on various factors, including model scores on different datasets. For the purpose of the experiment, a model trained on each node dataset and then used to predict the class for other nodes dataset. 

 The ensembled machine learning model is a collection of local models trained on each IoT node's local data. The edge node will receive and ensemble these local models using ensemble techniques and then share the ensembled model, which will represent the general model, with IoT devices. The ensembled model will use majority voting to decide on the classification problem.

 When evaluating the performance of classification models for botnet detection in IoT networks, \textbf{accuracy} and \textbf{training time} are the primary metrics used for comparison. Accuracy measures the proportion of correctly classified instances out of the total instances, providing a clear indication of a model’s effectiveness in distinguishing between benign and malicious network traffic. This metric is particularly valuable for assessing the overall performance of lightweight machine learning models, such as Decision Trees, Support Vector Machines (SVM), k-Nearest Neighbors (KNN), and Logistic Regression, as it directly reflects their ability to correctly identify botnet activities. Alongside accuracy, training time is a critical measure, especially in resource-constrained IoT environments, where models must be efficient enough to train quickly and deploy without overburdening the device’s capabilities. While other metrics like precision, recall, F1-score, and ROC-AUC can provide additional insights into the models’ performance and help fine-tune or enhance them, accuracy and training time remain the primary benchmarks for comparing the suitability of these models for real-time botnet detection in IoT networks.
 
 The process of comparing these four models will start with constructing the machine learning model by training on each dataset out of the seven and calculating the time consumed to train the model on each dataset and its accuracy. Then the weighted average for accuracy and training time will be calculated for datasets for all seven IoT devices, and the final score will be calculated based on this formula:

\begin{multline}\label{score_formula}
    Score = (accuracy\_weight * normalized\_accuracy) \\
        + (training\_time\_weight * normalized\_training\_time)       
\end{multline}

 "Accuracy weight" and "Time weight" are assigned weights that determine a model's score based on its accuracy and training time. High weights mean that models with high accuracy or less training time will receive a high score. We use normalized accuracy and normalized training time to facilitate score calculation and mode. We tested all models on Google Colab to ensure that the reported training time accurately represents the actual time taken by the model. To select a balanced model, the accuracy and time weights are initially set equally at 50\%.

\section{Results, Discussion and Future Work} \label{RD}

The initial results show that training a Support Vector Machine (SVM) on large datasets is very time-consuming. For example, training SVM on the first node data from NaBIOT, which contains approximately 1,018,298 records and 116 features, takes over 30 minutes. In contrast, a decision tree classifier requires only 27.37 seconds for the same dataset. To enhance SVM's efficiency on such large datasets, utilizing LinearSVM is recommended. LinearSVM offers a balance between speed and accuracy, making it a viable alternative for complex data tasks. It approaches the performance of non-linear SVM but with significantly reduced training time, as outlined in  \cite{chauhan_problem_2019}. Given SVM's lengthy training time, it will be excluded from further comparisons. The analysis will now concentrate on decision tree, KNN, and Logistic Regression classifiers.

\begin{figure}
    \centering
\includegraphics[width=1\linewidth]{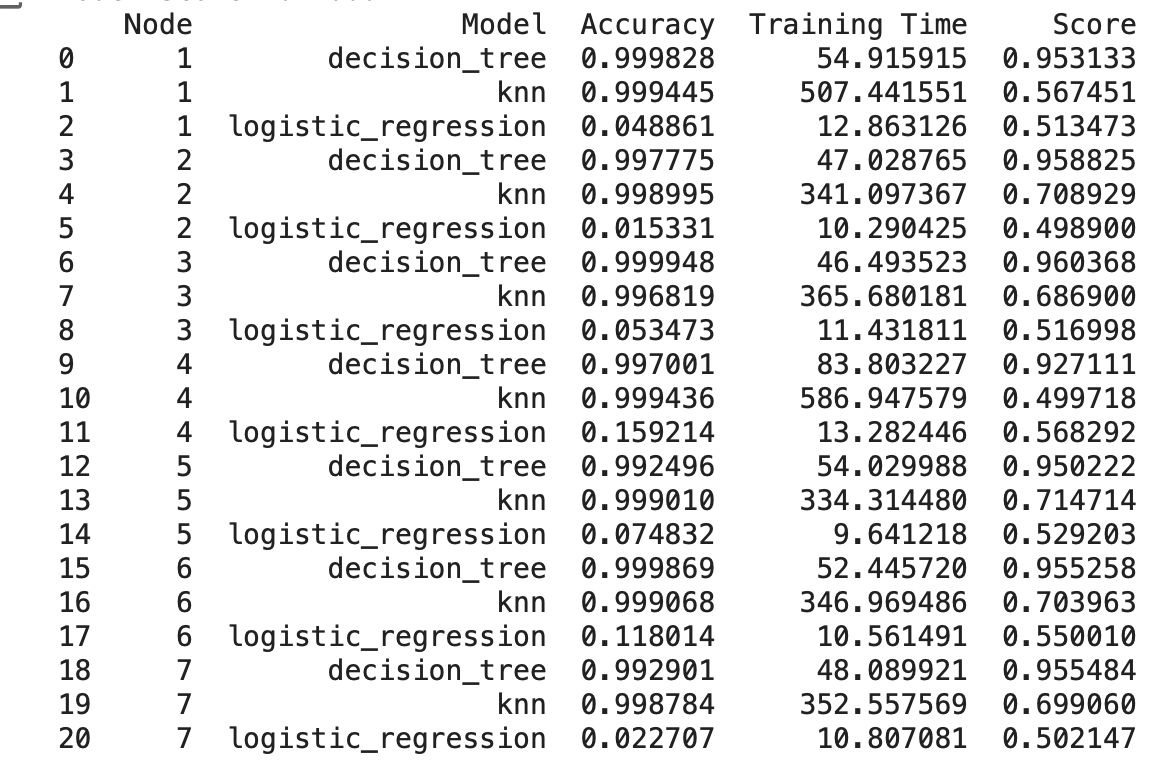}
    \caption{Statistics on model training results along with score by model per node}
    \label{fig:Statistics}
\end{figure}

Figure \ref{fig:Statistics} presents the accuracy, training time in seconds, and scores for each node after the decision tree, KNN, and logistic regression models' training and validation phases. The score is calculated using the formula described at \ref{score_formula}.

Figure \ref{fig:Statistics} illustrate that KNN and Decision tree outperform logistic regression models in accuracy. In addition, the decision tree model outperforms other models regarding score.

The analysis indicates that the KNN average model accuracy slightly outperforms the decision tree. At the same time, it is clear that logistic regression has very bad average accuracy across all nodes compared with the Decision tree and KNN, 

According to the results, the training time for Decision Tree is better than that of KNN and logistic regression. The average training times for Decision Tree (50 seconds), KNN (400 seconds), and Logistic Regression (11 seconds) classifiers, illustrating the relative computational efficiency of each model. 

\begin{figure}
    \centering
\includegraphics[width=1\linewidth]{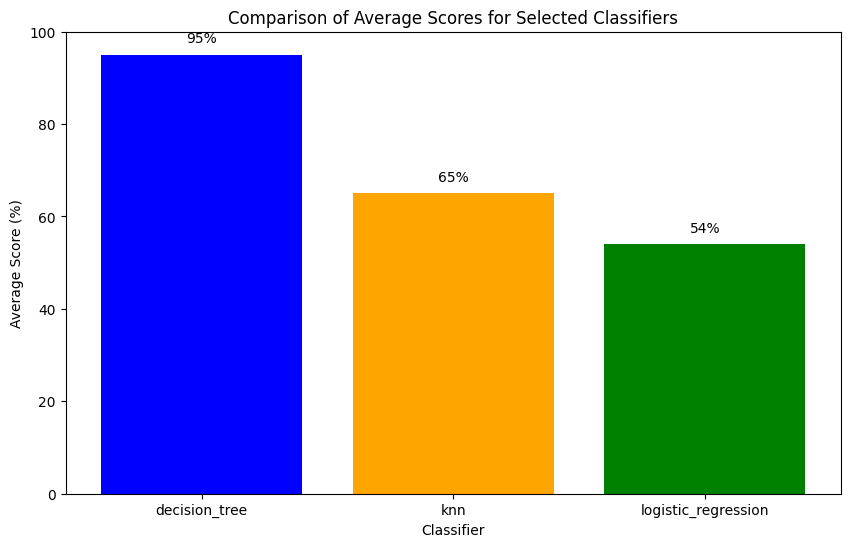}
    \caption{Average Score By Model}
    \label{fig:AVG_SCR_BY_MODEL}
\end{figure}

Based on the previous results, the average score of the Decision Tree across all nodes was expected to outperform that of KNN and logistic regression. Figure \ref{fig:AVG_SCR_BY_MODEL} displays the average score across all nodes for each model. The bar graph depicting the average scores for Decision Tree (95\%), KNN (65\%), and Logistic Regression (54\%) classifiers, highlighting their performance effectiveness.

Therefore, the Decision Tree can be justified as the most efficient algorithm out of the four machine learning models compared in this paper (Decision Tree, KNN, SVM, Logistic Regression) for botnet detection in local IoT devices, considering the balance between a high score and the lowest training time.

Selecting a lightweight machine learning model for botnet detection must consider its compatibility with federated learning. While efficient on local datasets, ML models l often struggle with generalization across different IoT nodes. Decision trees tend to perform better than KNN in terms of generalization across various datasets. Federated learning helps improve model performance across distributed datasets while maintaining data privacy.

The K-Nearest Neighbors (KNN) accuracy results indicate significant variability in model performance when applied to data from different IoT nodes. For example, a model trained on Node 1 data demonstrates nearly perfect accuracy (99.9\%) when tested on its own data but shows a drastic drop, such as 3.8\%, when applied to data from Node 2.

Table \ref{tab:KNN_accu_all}  presents the accuracy matrix for KNN, where each row corresponds to the training node and each column to the testing node. Diagonal values generally exhibit high accuracy, reflecting strong performance on native data. However, off-diagonal entries reveal substantial accuracy declines, confirming that KNN models tend to overfit local node characteristics and struggle to generalize to unseen data from other nodes.

\begin{table*}[ht]
\centering
\caption{Accuracy Matrix for KNN}
\[
\begin{array}{c|ccccccc}
 & \text{Node 1} & \text{Node 2} & \text{Node 3} & \text{Node 4} & \text{Node 5} & \text{Node 6} & \text{Node 7} \\
\hline
\text{Node 1} & 0.99944515 & 0.03805229 & 0.04217339 & 0.0483436  & 0.03946345 & 0.03920702 & 0.03795841 \\
\text{Node 2} & 0.60803223 & 0.99899507 & 0.15702921 & 0.16892772 & 0.11481057 & 0.12355253 & 0.14884124 \\
\text{Node 3} & 0.80573074 & 0.13246941 & 0.99681946 & 0.20622166 & 0.11832516 & 0.11880281 & 0.13112904 \\
\text{Node 4} & 0.73886426 & 0.01534199 & 0.05312402 & 0.99943569 & 0.0737522  & 0.11567217 & 0.02266974 \\
\text{Node 5} & 0.73768975 & 0.18179012 & 0.1787833  & 0.26772291 & 0.99900997 & 0.1879743  & 0.17668007 \\
\text{Node 6} & 0.72052484 & 0.01203289 & 0.01298641 & 0.13385736 & 0.03999831 & 0.99906798 & 0.00444744 \\
\text{Node 7} & 0.47177742 & 0.2015933  & 0.19423189 & 0.18687385 & 0.1637312  & 0.16343586 & 0.99878354 \\
\end{array}
\]
\label{tab:KNN_accu_all}
\end{table*}

In contrast, Decision Tree classifiers demonstrate more robust generalization across different nodes. Table \ref{tab:DT_acuu_all} illustrates that while there are variations, the accuracy drop is generally less severe compared to KNN. For instance, a model trained on Node 1 data achieves 99.99\% accuracy on the same node, but still maintains 75.99\% on Node 2 and 54.27\% on Node 5. This is an indication of better generalization than the KNN model.

\begin{table*}[ht]
\centering
\caption{Accuracy Matrix for Decision Tree}
\[
\begin{array}{c|ccccccc}
 & \text{Node 1} & \text{Node 2} & \text{Node 3} & \text{Node 4} & \text{Node 5} & \text{Node 6} & \text{Node 7} \\
\hline
\text{Node 1} & 0.99982814 & 0.75996799 & 0.75875378 & 0.83785043 & 0.54274745 & 0.52094478 & 0.74797197 \\
\text{Node 2} & 0.46303832 & 0.9977748  & 0.99395288 & 0.39156913 & 0.70406394 & 0.6967108  & 0.98342669 \\
\text{Node 3} & 0.57771988 & 0.99306117 & 0.99994786 & 0.51056953 & 0.60020525 & 0.59209622 & 0.98775895 \\
\text{Node 4} & 0.9540331  & 0.58516813 & 0.55645868 & 0.99700095 & 0.63081158 & 0.65885761 & 0.56656708 \\
\text{Node 5} & 0.80921498 & 0.57770291 & 0.55535562 & 0.82611268 & 0.99249632 & 0.99605803 & 0.57492601 \\
\text{Node 6} & 0.71062597 & 0.74806191 & 0.76814714 & 0.73545364 & 0.98888634 & 0.99986856 & 0.75560808 \\
\text{Node 7} & 0.41336426 & 0.9925922  & 0.99643129 & 0.38157712 & 0.52581798 & 0.53708547 & 0.99290105 \\
\end{array}
\]
\label{tab:DT_acuu_all}
\end{table*}

These patterns suggest that decision trees, while not immune to performance degradation when generalized, handle data variability across nodes more effectively than KNN. This indicates their suitability for federated machine learning in IoT networks, where adaptability across distributed data sources is essential. Incorporating decision trees into a federated learning framework enhances cross-device robustness while safeguarding data privacy through local training.

The final section of this project demonstrates ensemble techniques to simulate federated machine learning between different nodes. The table \ref{tab:ensemble_model_detail} illustrates the comparative effectiveness of using individual decision tree models versus an ensemble approach across seven different nodes, each trained on distinct datasets from various network devices but structured identically. Average accuracies per node highlight the performance of models trained individually on each node and applied to predict data across the network, ranging from 69.14\% to 81.52\%. Remarkably, the ensemble model, which aggregates predictions through a majority voting mechanism involving equal weights for each model, demonstrates superior accuracy, achieving values between 85.72\% and 98.97\%.

\begin{table}[h]
\centering
\label{tab:ensemble_model_detail}
\caption{Comparison of node-wise accuracies for the decision tree model using individual and ensemble techniques.}
\begin{tabular}{@{}ccc@{}}
\toprule
Node & Avg Accuracy per Node & Ensemble Accuracy \\ \midrule
1    & 73.83\%               & 85.72\%          \\
2    & 74.72\%               & 98.57\%          \\
3    & 75.16\%               & 98.82\%          \\
4    & 70.70\%               & 87.26\%          \\
5    & 76.17\%               & 97.88\%          \\
6    & 81.52\%               & 98.97\%          \\
7    & 69.14\%               & 97.77\%          \\ \bottomrule
\end{tabular}
\end{table}

This notable improvement underscores the substantial benefits of federated machine learning in enhancing botnet detection across a distributed network. By leveraging shared insights without compromising data privacy, federated learning enables nodes to detect novel and previously unseen patterns more effectively. The results also hint at potential further enhancements: adjusting the weights in the ensemble model based on individual model performances could likely yield even better outcomes. Such tailored weighting would optimize the overall accuracy by giving more influence to models demonstrating higher reliability, thus reinforcing the detection framework against botnet threats in diverse environments.

Future work may focus on defining a formal threat model and implementing defenses such as robust aggregation to enhance security. Applying statistical significance tests would strengthen the evaluation of model performance. Expanding the simulation setup to include more realistic conditions and device variability can improve practical relevance. Exploring deep learning baselines and optimizing communication overhead will further support the deployment of federated learning in real-world IoT environments.

\section{Conclusion} \label{conc}
This study introduced a botnet detection framework tailored for IoT environments, where devices often operate under strict resource constraints. Among the four lightweight supervised machine learning models evaluated, the Decision Tree model demonstrated the most effective balance between detection accuracy and training time, making it a strong candidate for real-time deployment. To enhance privacy and generalization, a federated learning setup was applied, allowing models to be trained locally on each device without sharing sensitive data. The ensemble approach used in this setup significantly improved detection performance, achieving up to 98.97\% accuracy across multiple nodes. These findings highlight the potential of combining lightweight models with federated learning to address key challenges in IoT security, offering a practical and scalable solution while laying the groundwork for future improvements in model aggregation and real-world deployment.

\bibliographystyle{apalike}
\bibliography{pubs2}
\end{document}